\documentclass[conference]{IEEEtran}
\usepackage{cite}
\usepackage{amsmath,amssymb,amsfonts}
\usepackage{algorithmic}
\usepackage{graphicx}
\usepackage{textcomp}
\usepackage{xcolor}
\usepackage{hyperref}
\usepackage{latexsym}
\usepackage{multirow} 
\usepackage{url}
\usepackage{float}
\usepackage{subfigure}
\usepackage{bm}
\usepackage{nicefrac}     
\usepackage{microtype}      
\usepackage{lipsum}
\usepackage[ruled,vlined]{algorithm2e}
\usepackage{booktabs}
\usepackage{epsfig}
\usepackage{epstopdf}
\usepackage{pifont}
\usepackage{bbding}

\def\BibTeX{{\rm B\kern-.05em{\sc i\kern-.025em b}\kern-.08em
    T\kern-.1667em\lower.7ex\hbox{E}\kern-.125emX}
}

\newcommand{\name}{GraphSC}
\makeatletter
\newcommand{\linebreakand}{%
  \end{@IEEEauthorhalign}
  \hfill\mbox{}\par
  \mbox{}\hfill\begin{@IEEEauthorhalign}
}
\def\thanks#1{\protected@xdef\@thanks{\@thanks
        \protect\footnotetext{#1}}}
\makeatother

\begin{document}

\title{Graph Self-Contrast Representation Learning\\
}

\author{\IEEEauthorblockN{Minjie Chen}
\IEEEauthorblockA{\textit{School of Data Science and
Engineering} \\
\textit{East China Normal University
}\\
Shanghai, China \\
\href{minjiechen@stu.ecnu.edu.cn}{minjiechen@stu.ecnu.edu.cn}
}
\and
\IEEEauthorblockN{Yao Cheng}
\IEEEauthorblockA{\textit{School of Data Science and
Engineering} \\
\textit{East China Normal University
}\\
Shanghai, China \\
\href{52215903009@stu.ecnu.edu.cn}{52215903009@stu.ecnu.edu.cn}
}
\and
\IEEEauthorblockN{Ye Wang}
\IEEEauthorblockA{\textit{School of Data Science and
Engineering} \\
\textit{East China Normal University
}\\
Shanghai, China \\
\href{yewang@stu.ecnu.edu.cn}{yewang@stu.ecnu.edu.cn}
}

\linebreakand

\IEEEauthorblockN{Xiang Li\IEEEauthorrefmark{1}} 
\IEEEauthorblockA{\textit{School of Data Science and
Engineering} \\
\textit{East China Normal University
}\\
Shanghai, China \\
\href{xiangli@dase.ecnu.edu.cn}{xiangli@dase.ecnu.edu.cn}}

\and

\IEEEauthorblockN{Ming Gao}
\IEEEauthorblockA{\textit{School of Data Science and
Engineering} \\
\textit{East China Normal University
}\\
Shanghai, China \\
\href{mgao@dase.ecnu.edu.cn}{mgao@dase.ecnu.edu.cn}}
}

\thanks{\IEEEauthorrefmark{1} Corresponding author.}
\maketitle

\begin{abstract}
Graph contrastive learning (GCL) has recently emerged as a promising approach for graph representation learning. 
Some existing methods adopt the 1-vs-$K$ scheme to construct one positive and $K$ negative samples for each graph, 
but it is difficult to set $K$.
For those methods that do not use negative samples, 
it is often necessary to add additional strategies to avoid model collapse,
which could only alleviate the problem to some extent.
All these drawbacks will undoubtedly have an adverse impact on the generalizability and efficiency of the model.
In this paper,
to address these issues,  
we propose a novel graph self-contrast framework \name, 
which only uses one positive and one negative sample, and chooses triplet loss as the objective. 
Specifically, self-contrast has two implications.
First,
\name~generates both positive and negative views of a graph sample from the graph itself via graph augmentation functions of various intensities, {and use them for self-contrast}. 
Second, \name~uses Hilbert-Schmidt Independence Criterion (HSIC) to factorize the representations into multiple factors and proposes a masked self-contrast mechanism to better separate positive and negative samples.
Further,
Since the triplet loss only optimizes the relative distance between the anchor and its positive/negative samples, it is difficult to ensure the absolute distance between the anchor and positive sample. Therefore, we explicitly reduced the absolute distance between the anchor and positive sample to accelerate convergence.
Finally,
we conduct extensive experiments to evaluate the performance of \name~against 19 other state-of-the-art methods in both unsupervised and transfer learning settings.
\end{abstract}


\section{Introduction}
Graph self-supervised learning (GSSL)~\cite{kipf2016variational,velickovic2019deep,you2020graph} has attracted significant attention in recent years. Compared with traditional semi-supervised and supervised graph learning~\cite{kipf2016semi,velivckovic2017graph,xu2018powerful},
GSSL seeks to employ supervision extracted from data itself,
which can effectively circumvent the need for costly annotated data.
In particular,
one of the main types of GSSL is graph contrastive learning (GCL)~\cite{sun2019infograph,you2020graph}, 
whose core idea is to minimize the distance between representations of different augmented views of the same graph (``positive pairs''), 
and maximize that of augmented views from different graphs (``negative pairs''). 
\begin{figure}[t]
  \centering
  \includegraphics[width=0.9\linewidth]{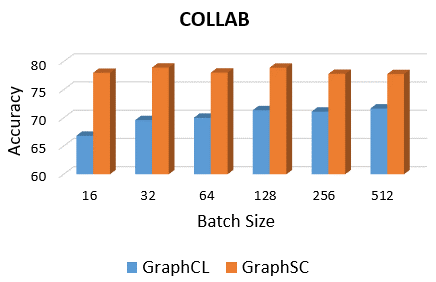}
  \caption{A toy example to show the influence of the number of negative samples $K$ on the model performance. In our experiments, we set $K = B-1$ in GraphCL and $K=1$ in our model \name, where $B$ is the batch size.}
  \label{Figure 1}
\end{figure}

According to whether negative samples are used by the model or not, 
most existing {graph-level} GCL methods fall into one of two classes. 
On the one hand, 
some approaches construct one positive sample and $K$ negative samples for each graph~\cite{you2020graph,xia2022simgrace,li2022let},
and formulate their objectives based on the normalized temperature-scaled cross entropy loss (NT-Xent)~\cite{sohn2016improved},
such as GraphCL~\cite{you2020graph}. 
However, 
these methods are easily affected by $K$ and an appropriate $K$ value is usually set empirically, which lacks theoretical supports.
When $K$ is small,
the model might not learn sufficient information to discriminate positive and negative samples;
otherwise,
there could lead to a large number of false-negative samples and slow convergence. 
In these methods,
for each graph in a batch,
other graphs in the same batch are considered as its negative samples,
i.e.,
$K = B -1$,
where $B$ is the batch size.
As shown in Fig. \ref{Figure 1}, 
the performance of 
GraphCL is significantly affected by $K$ on the COLLAB dataset.
In particular,
when $K$ is small,
the performance of GraphCL drops {drastically}.
On the other hand,
the second type of methods propose to not use negative samples.
However,
these methods could suffer from a degenerate solution \cite{zhu2021empirical}, 
where all outputs ``crash'' to an undesired constant.
To avoid such model collapse, 
additional strategies have to be applied, such as asymmetric dual encoders \cite{grill2020bootstrap,chen2021exploring}. 
Recently,
some studies \cite{li2022understanding} have showed that although these training strategies can avoid collapse to some extent, they may still cause collapse in partial dimensions of the representation, which leads to worse performance. 
The main reason for the model collapse is the complete non-use of negative samples.
Therefore, a research question arises: 
\emph{
To avoid the problem of $K$ selection and the  degenerate solution,
can we develop a {GCL} model that constructs only one positive sample and one negative sample for each graph?
}

Given one positive sample and one negative sample for each graph,
a straightforward framework is to use triplet loss as objective function. 
However, 
triplet loss is hard-to-train and mainly suffers from poor local optima and slow convergence, 
partially due to that the loss function employs only one negative example while not interacting with other negative classes per update~\cite{sohn2016improved}.
In short, there are two difficulties: 
one is to find a valid negative sample and the other is to solve the hard-to-train problem.

For the first difficulty,
hard negative sample mining~\cite{chuang2020debiased, robinson2020contrastive, wu2020conditional, kalantidis2020hard,schroff2015facenet} has been proposed.
However, most existing methods applied to graphs are node-level sampling, and very few is for graph-level sampling. 
Recently,
Cuco~\cite{chu2021cuco} proposes curriculum contrastive learning, 
which ranks negative samples from easy to hard and trains them in order.
CGC~\cite{Yang2022GeneratingCH} proposes to obtain reliable counterfactual negative samples by pre-training to help contrastive learning.
However, 
this introduces additional computational overhead 
that limits the performance of GCL. 
Inspired by the fact that 
some substances change their properties in response to external conditions, 
we propose a simple yet effective method to obtain negative samples from graph themselves.
For example, 
an enzymatic protein could become a non-enzymatic one after some perturbations. 
Since the non-enzymatic protein is directly generated from the enzymatic protein, 
they can share structural similarities to some degree,
which makes the negative sample discriminatively difficult, thereby achieving a similar effect to hard negative sampling.

To address the hard-to-train problem, 
we consider multiple facets of each graph to construct masked embedding vectors for its positive/negative samples.
Then the self-contrast is performed not only between the whole embedding vectors,
but also between masked embedding vectors corresponding to each facet.
The masked contrast can be used to provide more information and speed up the model convergence.
Further,
optimizing the triplet loss essentially maximizes the distance between positive and negative samples. This amplifies the margin between different classes but cannot ensure low-dimensional representations for each class compact. Therefore, we further shorten the absolute distance between anchor and positive sample, which can make each class more compact and make the distance between similar samples in the feature space closer.
It can also provide shortcuts for model convergence (we will show the experimental results in Section~\ref{sec:exp}).

In this paper, 
we study graph contrastive learning and propose a novel \textbf{Graph} \textbf{S}elf-\textbf{C}ontrast framework \name, 
which follows the pattern of generating positive and negative samples from the samples themselves and conducting self-contrast.
For each graph,
\name\ first generates one positive sample and one negative sample from the graph itself, 
and then self-contrasts the graph with its positive/negative samples as well as their masked embeddings.
Inspired by the assumption in~\cite{you2020graph} that the semantics of a graph will not change for a certain perturbation strength, 
we move forward and assume that the semantics of a graph will change under strong perturbations. 
Specifically, 
we propose to generate two different (positive and negative) views of a graph via graph augmentation functions of various intensities.
After that, 
the original graph and two generated views of the graph are fed into a shared GNN encoder,
after which sum pooling is used to derive graph-level representations. 
In particular,
we use the representation of the original graph as anchor, and the representations of views generated by weak and strong perturbations 
as a positive sample and a negative sample, respectively.
Further,
to implement masked self-contrast, 
we perform a division on the embeddings of positive/negative samples, and   
divide each representation vector into multiple factors by Hilbert-Schmidt Independence Criterion (HSIC)~\cite{gretton2005measuring}.
In addition to the contrast between the whole embedding vectors,
we mask each factor separately and perform masked self-contrast between corresponding representations.
Moreover,
we use Mean Square Error (MSE) loss/Barlow Twins loss (BT)~\cite{zbontar2021barlow} as a regularization to shorten the absolute distance between anchor and positive sample.
This leads to better convergence in implementation.
Finally,
we summarize the contributions as follows:
\begin{itemize}
\item 
We propose a novel graph self-contrast representation learning framework \name. 

\item
We present a simple yet effective method to construct negative samples from
graphs themselves in graph-level representation learning.

\item
We use triplet loss in graph contrastive learning and 
address the hard-to-train problem of triplet loss
by putting forward
a masked self-contrast mechanism and directly shortening the absolute distance for positive pairs.

\item
We conduct extensive experiments to evaluate the performance of \name\ in both unsupervised learning and transfer learning settings.
Experimental results 
show that \name~ 
performs favorably against other state-of-the-arts. 
\end{itemize} 

\section{RELATED WORK}
\subsection{Graph self-supervised learning}
Graph self-supervised learning~\cite{kipf2016variational,velickovic2019deep,you2020graph} 
aims to extract 
informative knowledge from graphs through pre-designed pretext tasks without relying on manual labels.
They can be used to alleviate the annotation bottleneck that is one of the main barriers for practical deployment of deep learning today. 
According to the objectives of pretext tasks, 
existing graph self-supervised learning methods 
can be broadly divided into four categories: 
(1) generation-based methods~\cite{kipf2016variational}, which aim to reconstruct the input graph data and 
use the input data as their supervision signals; 
(2) 
auxiliary-property-based methods~\cite{you2020does}, which attempt to obtain graph-related properties from the graph and further take them as supervision signals, such as pseudo labels of unlabeled data;
(3) contrast-based methods~\cite{velickovic2019deep,you2020graph}, which construct 
positive and negative pairs for contrast.
These methods
follow the core idea of
maximizing
the mutual information (MI)~\cite{hjelm2018learning}  between positive pairs 
and minimizing that between 
negative pairs.
(4) hybrid methods~\cite{zhang2020graph}, which integrate various pretext tasks together in a multi-task learning fashion. 
Our proposed method \name~  
is contrast-based
and we next introduce contrast-based methods in detail.
For a comprehensive survey on graph self-supervised learning,
see~\cite{liu2022graph}.

\subsection{Graph contrastive learning}
According to the contrast mode, 
graph contrastive learning can be mainly divided into three categories: 
node-node contrast, 
node-graph contrast and graph-graph contrast.

For 
node-node contrast, 
the representative model GRACE~\cite{zhu2020deep} first generates two contrastive views of a graph via graph augmentation, and then pulls close the representations of samples in positive pairs while pushing away that of samples in inter-view and intra-view negative pairs.
GCA~\cite{zhu2021graph} further introduces 
an adaptive augmentation by incorporating various priors for topological and semantic aspects of the graph, which results in a more competitive performance. GCC~\cite{qiu2020gcc} utilizes random walk 
as augmentations to extract the contextual information. 
BGRL~\cite{thakoor2021bootstrapped} maximizes the MI between node representations from online and target networks.

There also exist methods~\cite{velickovic2019deep,mavromatis2020graph,hassani2020contrastive,jiao2020sub,sun2019infograph} that are based on node-graph contrast.
For example,
DGI~\cite{velickovic2019deep}
learns both local and global semantic information in graphs by 
contrasting node-level embeddings with the graph-level representation.
After that,
GIC~\cite{mavromatis2020graph} seeks to 
additionally capture cluster-level information 
by first 
clustering nodes 
based on their embeddings, 
and then maximizing the MI between nodes in the same cluster. 
MVGRL \cite{hassani2020contrastive} first generates two graph views via graph diffusion
and subgraph sampling. 
Then it trains graph encoders 
by 
{contrasting} node embeddings in a view and the graph-level representation in another view. 
Further,
SUBG-CON~\cite{jiao2020sub} 
uses 
triplet loss as objective function.
For each node,
it first extracts 
the top-k most informative neighbors 
to form a subgraph. 
Then it pulls close the distance  
between 
the representations of 
the node and the subgraph,
and pushes away that between the representations of the node and a randomly selected subgraph.

The third type of methods are 
based on 
graph-graph contrast. 
The early model
GraphCL~\cite{you2020graph} 
designs four types of graph augmentation 
(node dropping, edge perturbation, attribute masking and subgraph extraction), 
and then 
adopts 
the NT-Xent loss to 
learn the graph-level representation.
Further, 
JOAO~\cite{you2021graph} proposes a unified bi-level optimization framework to automatically select data augmentations. 
AD-GCL~\cite{suresh2021adversarial} uses adversarial graph augmentation strategies that enables GNNs to avoid capturing redundant information during training. 
Inspired by
Invariant Rationale Discovery (IRD),
RGCL~\cite{li2022let} puts forward rationale-aware augmentations for graph contrastive learning to preserve the critical information in the graph.
There are also methods that do not need data augmentations. 
For example, 
SimGRACE~\cite{xia2022simgrace} feeds the original graph into a GNN encoder and achieves data augmentation through perturbation of the encoder.

\section{Preliminary}
In this section,
we introduce basic concepts used in this paper.
\subsection{Graph Neural Networks (GNNs)}
Let \begin{math} G=(\mathcal{V},\mathcal{E}) \end{math} denote an undirected graph, 
where \begin{math} \mathcal{V} = \{v_{1}, v_{2}, \cdots, v_{N}\} \end{math} is the node set
and \begin{math} \mathcal{E} \subseteq \mathcal{V} \times \mathcal{V} \end{math} represents the edge set.
We use \begin{math} X \in \mathbb{R}^{N \times F} \end{math} to denote the node feature matrix, 
where $F$ is the dimension of node 
features. 
Generally, 
given a GNN model $f(\cdot)$,
message propagation in the $l$-th layer can be divided into two operations: one is to aggregate information from a node's neighbors while the other is to update a node's embedding. 
Taking node $v_i$ as an example,
we formally define these two operations as:
\begin{equation}
  a_{{i}}^{(l)} = \texttt{AGGREGATE}^{(l)}\{{h_{{j}}^{(l-1)}, \forall v_{j} \in \mathcal{N}(v_{i})}\},
\end{equation}
\begin{equation}
  h_{{i}}^{(l)} = \texttt{COMBINE} \{ h_{{i}}^{(l-1)}, a_{{i}}^{(l)} \},
\end{equation}
where \begin{math} h_{{i}}^{(l)} \end{math} is the embedding of node $v_{i}$ in the $l$-th layer 
and 
$\mathcal{N}(v_{i})$ is a set of nodes adjacent to $v_{i}$. 
\texttt{AGGREGATE}$^{(l)}(\cdot)$ and \texttt{COMBINE}$^{(l)}(\cdot)$ are two functions in each GNN layer. 
After $L$ propagation layers, 
the output embedding for $G$ is summarized on node embeddings 
via the \texttt{READOUT} function,
which is formulated as:
\begin{equation}
  f(G) = \texttt{READOUT}\{ h_{{i}}^{(L)}, \forall v_{i} \in \mathcal{V}\}.
\end{equation}

\subsection{Hilbert-Schmidt Independence Criterion}
The Hilbert-Schmidt Independence Criterion (HSIC)~\cite{gretton2005measuring} is a kernel-based measure of dependence between probability distributions.
Let $\mathcal{F}$ be a Hilbert space of real-value functions from a set $\mathcal{X}$ to $\mathbb{R}$. 
We say $\mathcal{F}$ is a  Reproducing Kernel Hilbert Space (RKHS) if 
$\forall x \in \mathcal{X}$, 
the Dirac evaluation operator $\delta_{x}$ : $\mathcal{F} \rightarrow \mathbb{R}$, which maps $f \in \mathcal{F}$ to $f(x) \in \mathbb{R}$, is a 
bounded linear functional.
In RHKS,
$\forall x \in \mathcal{X}$,
there is a mapping $\phi(x) \in \mathcal{F}$
and 
there also exists a unique
definite kernel $u$ : $\mathcal{X} \times \mathcal{X} \rightarrow \mathbb{R}$, such that $\langle\phi(x), \phi(x^{'})\rangle_{\mathcal{F}} 
=u(x,x{'})
$.

Assume that we have two separable RKHSs $\mathcal{F}$, 
$\mathcal{G}$ 
and a joint measure $p_{xy}$ over ($\mathcal{X} \times \mathcal{Y}$, $\Gamma \times \Lambda$), 
where 
$\Gamma$ is the Borel sets on set $\mathcal{X}$ and $\Lambda$ is the Borel sets on set $\mathcal{Y}$.
Then the Hilbert-Schmidt Independence Criterion (HSIC) 
is defined as 
the squared Hilbert-Schmidt norm of the associated cross-covariance operator $C_{xy}$:
\begin{equation}
HSIC(p_{xy},\mathcal{F},\mathcal{G}):=||C_{xy}||_{HS}^{2},
\end{equation}
where the Hilbert-Schmidt norm is $||A||_{HS}=\sqrt{\sum_{i,j}a_{ij}^{2}}$,
and
the cross-covariance operator is given as follows:
\begin{equation}
  C_{xy}:=\textbf{E}_{x,y}[(\phi(x)-\mu_{x})\otimes(\varphi(y)-\mu_{y})],
\end{equation}
Here, 
$\otimes$ is tensor product, and $\phi(\cdot)$, $\varphi(\cdot)$ are functions that map $x\in \mathcal{X}$ and $y\in \mathcal{Y}$ to RKHSs $\mathcal{F}$ and $\mathcal{G}$ w.r.t. the kernel functions $u(x,y)=<\phi(x),\phi(y)>$ and $s(x,y)=<\varphi(x),\varphi(y)>$, respectively.
Accordingly, 
given i.i.d. $m$ samples $(X,Y)=\{(x_{1},y_{1}),\cdots, (x_{m},y_{m})\}$ drawn from the joint distribution of $p_{xy}$,
the empirical version of HSIC
is given as:
\begin{equation}
  HSIC(X,Y)=(m-1)^{-2}tr(UHSH),
\end{equation}
where $U,S,H \in \mathbb{R}^{m \times m}$, $U_{ij} :=u(x_{i},x_{j})$, $S_{ij} :=s(y_{i},y_{j})$ and $H_{ij} :={I} - m^{-1}$.

\begin{figure*}[t]
  \centering
  \includegraphics[width=0.9\textwidth]{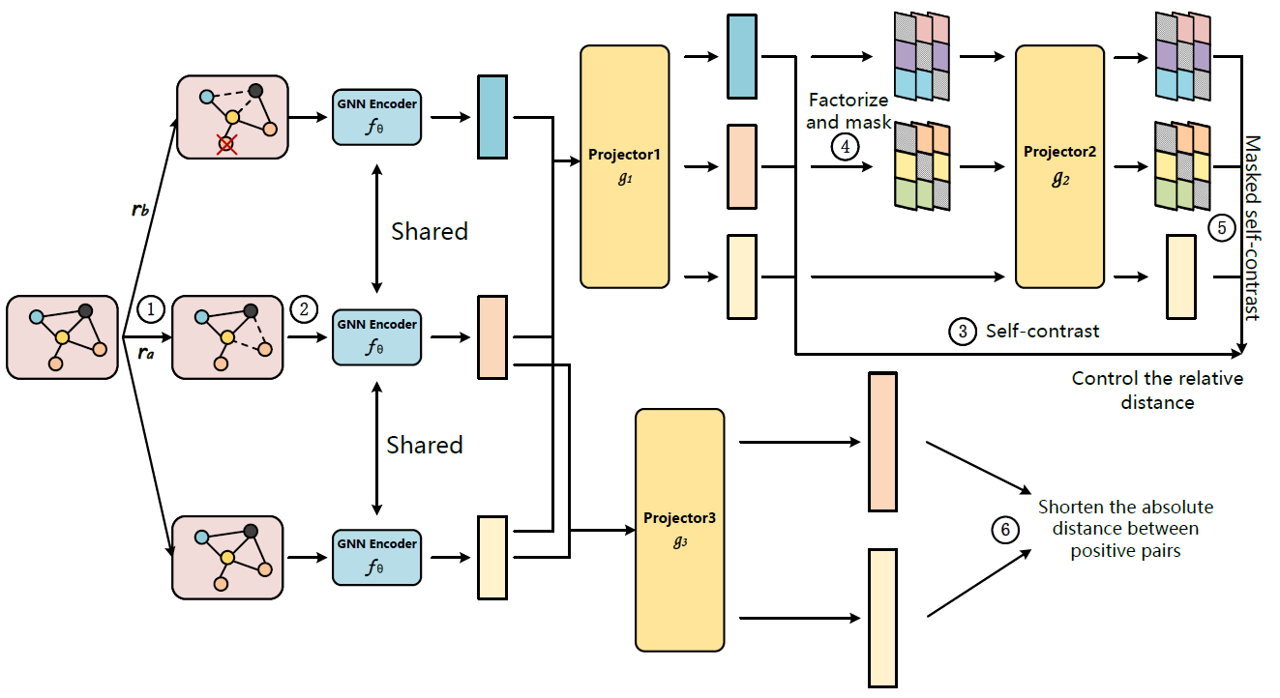}
  \caption{The overall framework of \name. For details of each step, see Section \ref{sec method}.}
  \label{Figure 2}
\end{figure*}

\section{METHOD}

\label{sec method}
In this section, we introduce the \name\ framework.
Given a graph, 
\name~ first generates two augmented views as positive and negative samples via weak and strong perturbation, respectively (Step \ding{172}). 
Then the graph and its augmented views are fed into a GNN encoder with shared parameters to obtain corresponding graph-level representations (Step \ding{173}). 
After these representations have been mapped, they are self-contrasted (Step \ding{174}). 
At the same time, considering each graph has multi-facet features, 
\name~first factorizes the representations of positive and negative samples by using HSIC,
and then masks each factor sequentially to generate multiple masked views (Step \ding{175}).
The representations of anchors and the masked representations of positive and negative samples are contrasted after a projection head (Step \ding{176}). 
Finally, \name~shortens the absolute distance between an anchor and its positive sample (Step \ding{177}).
The overall framework of \name\ is given in Fig. \ref{Figure 2}.

\subsection{Data augmentation}
\label{DA}
To construct positive and negative pairs,
most existing graph contrastive learning methods~\cite{you2020graph,suresh2021adversarial} first perform data augmentations on graphs, such as node dropping and edge perturbation.
After that,
for each graph,
its augmented views of graphs form positive samples
while that of other graphs in the same mini-batch are considered as negative samples.
Despite the success,
these methods can be easily affected by the number of negative samples $K$.
To mitigate the influence of 
$K$ on the model performance and reduce the number of false negative samples selected,
using hard negative samples
could be a feasible solution.
However, 
general hard negative sample mining strategies are either not suitable for graph data~\cite{xia2022progcl} or computationally costly~\cite{chu2021cuco}.

Inspired by the assumption in GraphCL~\cite{you2020graph} 
that the semantics of a graph will not change for a certain perturbation strength, 
we further assume that the semantics of a graph will change under strong perturbations. 
Based on these two assumptions,
we generate positive and negative pairs from graphs themselves via graph augmentation functions of various intensities.
Specifically,
given a graph $G$,
the generated view of the graph with weak perturbation is considered as a positive sample $G^{+}$,
while that generated by strong perturbation is taken as negative sample $G^{-}$.
Formally,
for any graph augmentation function $\mathcal{A}(\cdot)$ and two different perturbation rates $ r_{a}$, $ r_{b} $,
where $ (r_{a}<r_{b}) $ ,
we have

\begin{equation}
  G^{+}=\mathcal{A}(G;r_{a}),G^{-}=\mathcal{A}(G;r_{b}).
\end{equation}

In this way,
since negative samples are directly constructed from graphs themselves,
they can share similarities with the original graphs to some degree.
Therefore,
these negative samples can play the role of hard negative,
which can boost the model performance.

\subsection{Model architecture}
\textbf{Self-contrast:}
Given a graph $G_i$ and its augmentations $G_{i}^{+}$ and $G_{i}^{-}$,
we first feed them into a shared GNN encoder $f(\cdot;\theta)$ to learn graph-level representation vectors \begin{math} h_{i}, h_{i}^{+}, h_{i}^{-} \end{math}, respectively.
We denote
\begin{equation}
  h_{i}=f(G_{i}), h_{i}^{+}=f(G_{i}^{+}), h_{i}^{-}=f(G_{i}^{-}).
\end{equation}
After that,
as suggested in~\cite{chen2020simple}, 
the projection head,
a non-linear transformation, 
can be used to map these representations to another latent space,
which can enhance the model performance.
Therefore,
we further define 
a projection head $g_{1}(\cdot;\phi_{1})$
and derive:
\begin{equation}
  y_{i} = g_1(h_{i}),y_{i}^{+} = g_1(h_{i}^{+}),y_{i}^{-} = g_1(h_{i}^{-}),
\end{equation}
where \begin{math} y_{i},y_{i}^{+},y_{i}^{-} \in \mathbb{R}^{d} \end{math}.
These vectors 
characterize the overall feature information of samples,
which can be used for contrast.

\textbf{Masked self-contrast:}
Further,
since each graph has multi-facet features, 
in order to pull apart positive pairs from negative pairs,
we can perform contrastive learning from masked views.
Specifically, 
we first factorize the representations $y_{i}^{+},y_{i}^{-}$ of positive and negative samples into $n$ independent factors \begin{math} y_{i}^{+}=[c_{i1}^{+},c_{i2}^{+}, \cdots ,c_{in}^{+}] \end{math} and \begin{math} y_{i}^{-}=[c_{i1}^{-},c_{i2}^{-}, \cdots ,c_{in}^{-}] \end{math}, respectively. 
Then we sequentially mask each factor and generate $n$ views.
For the $m$-th masked view,
the corresponding embeddings of positive and negative samples are denoted as:  
\begin{equation}
\begin{split}
  & y_{im}^+ = [c_{i1}^{+}, \cdots ,c_{i(m-1)}^{+}, c_{0} ,c_{i(m+1)}^{+}, \cdots ,c_{in}^{+}] \text{ and} \\
  & y_{im}^- = [c_{i1}^{-}, \cdots ,c_{i(m-1)}^{-}, c_{0} ,c_{i(m+1)}^{-}, \cdots ,c_{in}^{-}], \\
\end{split}
\end{equation}
where 
$c_{0} \in \mathbb{R}^{d_{c}} $ is an all-zero vector and \begin{math} d_{c}=d/n \end{math}. 
We further feed
\begin{math} y_{i} \end{math}, \begin{math} y_{im}^{+} \end{math}, \begin{math} y_{im}^{-} \end{math} into the second projection head \begin{math} g_2(\cdot;\phi_{2}) \end{math} to obtain their projected embeddings:
\begin{equation}
  q_{i}^{} = g_{2}^{}(y_{i}^{}), q_{im}^{+} = g_{2}^{}(y_{im}^{+}), q_{im}^{-} = g_{2}^{}(y_{im}^{-}).
\end{equation}
These projected embeddings can be used for masked contrast across various views.

\textbf{Shorten the absolute distance:}
Finally,
since triplet loss can only capture the relative distance between an anchor and its positive/negative sample,
we explicitly shorten the absolute distance between an anchor and its positive distance.
To achieve the goal,
we introduce the third
projection head \begin{math} g_3(\cdot ; \phi_{3}) \end{math} 
and generate
\begin{equation}
  z_{i}^{} = g_3(h_{i}^{}),z_{i}^{+} = g_3(h_{i}^{+}),
\end{equation}
where \begin{math} z_{i},z_{i}^{+} \in \mathbb{R}^{d_{h}}\end{math}.
Then we explicitly pull close $ z_{i}$ and $ z_{i}^+$.

\subsection{Contrastive loss}

In the training process, 
we randomly select $B$ graphs from the whole dataset of \begin{math} N \end{math} graphs as a mini-batch.
For each graph $G_i$ in the mini-batch,
we apply two different strengths of perturbation,
which generates a triple $(G_{i},G_{i}^{+},G_{i}^{-})$.
Further,
we derive 
the corresponding representations $(h_{i},h_{i}^{+},h_{i}^{-})$ through a shared GNN encoder.

\textbf{Self-contrast:} To implement self-contrast between complete representations, 
we put $(h_{i},h_{i}^{+},h_{i}^{-})$ into a projection head to get $(y_{i},y_{i}^{+},y_{i}^{-})$, and use triplet margin loss to enlarge the relative distance between positive and negative sample pairs:
\begin{equation}
  \mathcal{L}_{se} = \frac{1}{B}\sum_{i=1}^{B}\max( ||y_{i}-y_{i}^{+}||^{2} - ||y_{i}-y_{i}^{-}||^{2} + \epsilon, 0),
  \label{eq:13}
\end{equation}
Note that $\epsilon$ is the margin.

\textbf{Masked self-contrast:}For masked self-contrast,
we first use Hilbert-Schmidt Independence Criterion (HSIC) \cite{gretton2005measuring}
to factorize $y_i^+$ and $y_i^-$ into $n$ factors:
\begin{equation}
  \mathcal{L}_{fa}=\frac{1}{B}\sum_{i=1}^{B} \sum_{j\not=k}^{n} [
  \texttt{HSIC}(c_{ij}^{+},c_{ik}^{+}) + \texttt{HSIC}(c_{ij}^{-},c_{ik}^{-})].
\end{equation}

This process
ensures that factors are as independent as possible from each other, which helps reduce the dependence between multiple partial representations.
After that,
we sequentially mask each factor to generate a set of masked representations $\{y_{im}^+\}_{m=1}^n$ and $\{y_{im}^-\}_{m=1}^n$, respectively. 
When a factor is masked,
it is still expected that the positive sample can be close to the anchor while the negative sample is distant.
Therefore,
we formulate the masked contrastive loss as:
\begin{equation}
\resizebox{0.42\textwidth}{!} {
$\mathcal{L}_{ma}=\frac{1}{B}\sum_{i=1}^{B}\sum_{m=1}^{n}w_{im} \cdot \max  ( \Vert  q_{i}^{} - q_{im}^{+} \Vert^{2} - \Vert q_{i}^{} - q_{im}^{-} \Vert^{2} + \epsilon, 0),$
}
\label{eq:15}
\end{equation}
where
\begin{equation}
  w_{im}= \left (1 - \frac{exp(e_{im})}{\sum_{i=1}^{n} exp\left(e_{im}\right )} \right) \cdot \frac{1}{n-1}, 
\end{equation}
\begin{equation}
  e_{im}= q_{i}^{} \cdot (q_{im}^{+} - q_{im}^{-})^{T}. 
\end{equation}
Here, 
we introduce 
the weight $w_{im}$ for the $m$-th factor and $\sum_{m=1}^n w_{im} = 1$.
For the factor that leads to a small relative distance,
our model will assign a large weight and thus pay more attention to the corresponding optimization process.

\textbf{Shorten the absolute distance:} 
To shorten the absolute distance between positive pairs, which can make each class more compact,
we propose two models \name~ and \name-MSE, which use Barlow Twins loss and MSE loss as regularization terms, respectively.

\textbf{(1) \name:}
\name~ pulls close the representations of the anchor and the positive sample after the third projection head $g_{3}$, and we formulate the objective function $\mathcal{L}_{ab}$ as:
\begin{equation}
  \mathcal{L}_{ab}=\frac{1}{B}\sum_{i}(1-C_{ii}^{})^{2}+ \beta \sum_{i} \sum_{j\neq{i}}C_{ij}^{2},  
  \label{eq:18}
\end{equation}
\begin{equation}
  C_{ij}=\frac{\sum_{b=1}^{B}z_{b,i}^{}z_{b,j}^{+}}{\sqrt{\sum_{b=1}^{B} (z_{b,i}^{})^{2}} \sqrt{\sum_{b=1}^{B} (z_{b,j}^{+})^{2}}}. 
\end{equation}
Here,
Barlow Twins loss can additionally reduce the redundancy between components of embedding vectors,
which can also boost the model performance.

\textbf{(2) \name-MSE:}
\name-MSE no longer structurally needs the third projection head $g_{3}$, and the regularization term $\mathcal{L}_{ab}$ can be written as:
\begin{equation}
  \mathcal{L}_{ab} = \frac{1}{B}\sum_{i=1}^{B}||y_{i}-y_{i}^{+}||^{2},
  \label{eq:21}
\end{equation}
Note that the
MSE loss is a widely used distance measure, and we can use \name-MSE to verify the necessity and effectiveness of shortening the absolute distance between the anchor and the positive sample.
Finally, our objective function
is summarized as: 
\begin{equation}
  \min \mathcal{L}=\underbrace{\mathcal{L}_{se} + \lambda_{1} \cdot \mathcal{L}_{ma} + \lambda_{2} \cdot \mathcal{L}_{fa}}_{relative~ term} + \underbrace{\lambda_{3} \cdot \mathcal{L}_{ab}}_{absolute~ term}, 
  \label{eq:20}
\end{equation}
where $\lambda_1$, $\lambda_2$ and $\lambda_3$ are hyper-parameters that are used to balance the term importance.

\section{EXPERIMENTS}

\label{sec:exp}
In this section, we conduct 
experiments on multiple benchmark datasets to evaluate the performance of \name~through answering the following research questions.
\begin{itemize}
\item \textbf{RQ1. (Generalizability)} 
{Does \name\ outperform other competitors in unsupervised settings? }
\item \textbf{RQ2.(Transferability)} Can GNNs pre-train with GraphSC show better transferability than competitors?

\item \textbf{RQ3. (Effectiveness)}
Are the individual components of GraphSC really valid for the model ?

\item \textbf{RQ4. (Convergence)}
What is the effect of $\mathcal{L}_{ma}$, $\mathcal{L}_{ab}$ and the proposed negative sample generation strategy on the convergence of the model?

\item \textbf{RQ5. (Hyperparameters Sensitivity)} Is the proposed GraphSC sensitive to hyperparameters like perturbation intensity $r_{a}$, $r_{b}$ and the term weight $\lambda_1$, $\lambda_2$, $\lambda_3$?

\begin{table}[!b]
  \caption{Datasets statistics for unsupervised learning.}
  \footnotesize
  \setlength{\tabcolsep}{2pt}
  \label{Table 1}
  \begin{tabular}{c|c|c|c|c}
    \toprule
    Dataset & Category & Graph Num. & Avg. Node & Avg. Edge\\
    \midrule
    NCI1 & Biochemical Molecules & 4110 & 29.87 & 32.30 \\
    PROTEINS & Biochemical Molecules & 1113 & 39.06 & 72.82 \\
    DD & Biochemical Molecules & 1178 & 284.32 & 715.66\\
    MUTAG & Biochemical Molecules & 188 & 17.93 & 19.79\\
    \midrule
    COLLAB & Social Networks & 5000 & 74.49 & 2457.78\\
    RDT-B & Social Networks & 2000 & 429.63 & 497.75\\
    RDB-M & Social Networks & 4999 & 508.52 & 594.87\\ 
    IMDB-B & Social Networks & 1000 & 19.77 & 96.53\\
    \bottomrule
\end{tabular}
\end{table}

\end{itemize}

\subsection{Experimental Setup}

\textbf{Datasets}: 
For unsupervised learning,
we use 8 datasets from the benchmark TU dataset \cite{morris2020tudataset},
including graph data for various biochemical molecules (i.e., NCI1, PROTEINS, DD, MUTAG) and social networks (i.e., COLLAB, REDDIT-BINARY, REEDIT-MULTI-5K and IMDB-BINARY). 
For transfer learning, we perform pre-training on ZINC-2M which samples 2 million unlabeled molecules from ZINC15~\cite{sterling2015zinc} and fine-tune the model with 8 datasets including BBBP, Tox21, ToxCast, SIDER, ClinTox, MUV, HIV and BACE. 
More details can be seen in the Table \ref{Table 1} and Table \ref{Table 2}.

\begin{table*}[!t]
  \caption{Datasets statistics for transfer learning.}
  \label{Table 2}
  \resizebox{\linewidth}{!}{
  \begin{tabular}{c|c|c|c|c|c}
    \toprule
    Datasets & Category & Utilization & Graph Num. & Avg. Node  & Avg.Degree\\
    \midrule
    ZINC-2M & Biochemical Molecules & PRE-TRAINING & 2,000,000 & 26.62 &  57.72\\
    \midrule
    BBBP & Biochemical Molecules & FINETUNING & 2,039 & 24.06 & 51.90\\
    TOX21 & Biochemical Molecules & FINETUNING & 7,831 & 18.57 & 38.58\\
    TOXCAST & Biochemical Molecules & FINETUNING & 8,576 & 18.78 & 38.52\\
    SIDER & Biochemical Molecules & FINETUNING & 1,427 & 33.64 & 70.71\\
    CLINTOX & Biochemical Molecules & FINETUNING & 1,477 & 26.15 & 55.76\\
    MUV & Biochemical Molecules & FINETUNING & 93,087 & 24.23 & 52.55\\
    HIV & Biochemical Molecules & FINETUNING & 41,127 & 25.51 & 54.93\\
    BACE & Biochemical Molecules & FINETUNING & 1,513 & 34.08 & 73.71\\

  \bottomrule
\end{tabular}
}
\end{table*}

\textbf{Baselines}:
For unsupervised learning, 
we compare \name\ with 
three kernel-based methods including graphlet kernel (GL) \cite{shervashidze2009efficient}, Weisfeiler-Lehman kernel (WL) \cite{shervashidze2011weisfeiler}, and deep graph kernel (DGK) \cite{yanardag2015deep}. Furthermore, we compare \name~with other state-of-the-art 
methods: sub2vec \cite{adhikari2018sub2vec}, graph2vec \cite{narayanan2017graph2vec}, InfoGraph \cite{sun2019infograph}, GraphCL \cite{you2020graph}, {JOAO(v2)} \cite{you2021graph}, AD-GCL~\cite{suresh2021adversarial}, SimGRACE \cite{xia2022simgrace}, RGCL \cite{li2022let} and LaGraph \cite{xie2022self}.
We also take \name-MSE as our baseline.
For transfer learning, we adopt DGI \cite{velickovic2019deep}, EdgePred \cite{hu2019strategies}, AttrMasking \cite{hu2019strategies}, ContextPred \cite{hu2019strategies}, GraphCL \cite{you2020graph}, JOAO(v2) \cite{you2021graph}, AD-GCL~\cite{suresh2021adversarial}, SimGRACE \cite{xia2022simgrace}, GraphLoG\cite{Xu2021SelfsupervisedGR} and RGCL \cite{li2022let}, which are the state-of-the-art pre-training paradigms in this area, as our baselines.

\textbf{Evaluation Protocols}: Following the settings of previous works \cite{hu2019strategies, you2020graph,li2022let}, we evaluate the performance and generalizability of the learned representations on both unsupervised and transfer learning settings.
In unsupervised setting, we train \name~ using the whole dataset to learn graph representations and feed them into a downstream SVM classifier with 10-fold cross-validation, report the mean accuracy with standard deviation after 5 runs. 
For transfer learning, we pre-train and fine-tune GNN encoder in different datasets to evaluate the transferability of the pre-training scheme. 
The fine-tuning procedure is repeated for 10 times with different random seeds and we evaluate the mean and standard deviation of AUROC scores on each downstream dataset, which is consistent with our baselines.

\textbf{Implementation details}:
We implement \name~ using PyTorch. The model is initialized by Xavier initialization \cite{glorot2010understanding} and trained by Adam~\cite{kingma2014adam}. 
As suggested in~\cite{schroff2015facenet},
we set $\epsilon$ in  (\ref{eq:13}) and (\ref{eq:15}) to 0.2. Similarly, 
we set $\beta$ in (\ref{eq:18}) to 0.013 according to~\cite{zbontar2021barlow}. 
For other hyperparameters,
we fine-tune them by grid search.
For unsupervised learning,
we first fine-tune learning rate from $\{0.001, 0.005, 0.01\}$. For the augmentation functions $A(\cdot)$, we choose from four augmentations and some of their combinations, which are in line with GraphCL~\cite{you2020graph}. 
For the perturbation rates $r_{a}$ and $r_{b}$, we fine-tune them from \{0.05, 0.1, 0.15, 0.2\} and \{0.15,0.2,0.25,0.3,0.35,0.4\} respectively. In addition, we fine-tune $\lambda_1$, $\lambda_2$ and $\lambda_3$ from \{0.001, 0.01, 0.1, 1, 10, 100 \}. 
In transfer learning, we pre-trained the GNN encoder on the ZINC-2M dataset, and we set learning rate to 0.001, the number of epochs to 80, $r_{a}$ to 0.1, $r_{b}$ to 0.25, $\lambda_1$ to 1, $\lambda_2$ to 0.01 and $\lambda_3$ to 0.01. In addition, we use the combination of subgraph and node dropping as augmentation function, which is the same as GraphCL~\cite{you2020graph}. In the process of fine-tuning, we adjust the two hyperparameters learning rate and epoch, and the grid search range is \{0.0001, 0.0005, 0.001\} and \{20, 40, 60, 80, 100\} respectively.
Since most results of baselines are publicly available, 
we directly report these results from their original papers.
For the results of AD-GCL and GraphLoG,
we report these results from RGCL \cite{li2022let}.
For fairness,
we run all the experiments on a server with 128G memory and a single NVIDIA 2080Ti GPU. 
We provide our code and data here:
\url{https://anonymous.4open.science/r/GraphSC-8360}.

\subsection{Unsupervised learning (RQ1)}
For unsupervised representation learning,
we take the one-hot representations of node labels and degrees as node feature vectors for molecular datasets and social network datasets, respectively.
We summarize the experimental results in Table~\ref{Table 3}.
From the table, 
we see that
\name~ranks first on 5 out of 8 datasets and has competitive results on the other three. 
For example, 
the accuracie of \name~on the COLLAB datasets is 78.90\% , which is $>1.2\%$ higher than the runner-up. Moreover, \name~ also leads in the other four datasets (i.e. NCI1, PROTEINS, RED-B and RED-M5K) by 0.3\%-0.8\%.
Further, 
the average ranking of \name~ across all the datasets is 1.5, much better than the runner-up's, which is 2.8. 
We also notice that \name-MSE, which utilizes the MSE loss to minimize the absolute distance between anchor and positive sample, 
is the runner-up among all the methods.
This further verifies the effectiveness of our proposed self-contrast framework, 
which is not simply originated from the Barlow Twin loss regularization. 



\begin{table*}[t]
  \caption{Unsupervised representation learning 
  on TU datasets. 
  All the baselines are compared in the same experiment setting w.r.t. the classification accuracy(\%). \underline{Bold} indicates the best performance on each dataset. A.R. denotes average rank. - indicates that results are not available in published papers. Results for SOTA methods are as published.}
  \setlength{\tabcolsep}{4pt}
  \label{Table 3}
  \begin{tabular}{ccccc|cccc|c}
    \toprule
    Dataset & NCI1 & PROTEINS & DD & MUTAG & COLLAB & RDT-B & RDT-M5K & IMDB-B & A.R.\\
    \midrule
    GL & - & - & - & $81.66\pm2.11$ & - & $77.34\pm0.18$ & $41.01\pm0.17$ & $65.87\pm0.98$ & 13.3\\
    WL & $80.01\pm0.50$ & $72.92\pm0.56$ & - & $80.72\pm3.00$ & - & $68.82\pm0.41$ & $46.06\pm0.21$ & $72.30\pm3.44$ & 7.9 \\
    DGK & $80.31\pm0.46$ & $73.30\pm0.82$ & - & $87.44\pm2.72$ & - & $78.04\pm0.39$ & $41.27\pm0.18$ & $72.30\pm3.44$ & 6.4\\
    \midrule
    sub2vec & $52.84\pm1.47$ & $53.03\pm0.55$ & - & $61.05\pm15.80$ & - & $71.48\pm0.41$ & $36.68\pm0.42$ & $55.26\pm1.54$ & 14.5\\
    graph2vec & $73.22\pm1.81$ & $73.30\pm2.05$ & - & $83.15\pm9.25$ & - & $75.78\pm1.03$ & $47.86\pm0.26$ & $71.10\pm0.57$ & 11.5\\
    InfoGraph & $76.20\pm1.06$ & $74.44\pm0.31$ & $72.85\pm1.78$ & $89.01\pm1.13$ & $70.65\pm1.13$ & $82.50\pm1.42$ & $53.46\pm1.03$ & $73.03\pm0.87$ & 7.9\\
    GraphCL & $77.87\pm0.41$ & $74.39\pm0.45$ & $78.62\pm0.40$ & $86.80\pm1.34$ & $71.36\pm1.15$ & $89.53\pm0.84$ & $55.99\pm0.28$ & $71.14\pm0.44$ & 7.8\\
    JOAO & $78.07\pm0.47$ & $74.55\pm0.41$ & $77.32\pm0.54$ & $87.35\pm1.02$ & $69.50\pm0.36$ & $85.29\pm1.35$ & $55.74\pm0.63$ & $70.21\pm3.08$ & 8.9\\
    JOAOv2 & $78.36\pm0.53$ & $74.07\pm1.10$ & $77.40\pm1.15$ & $87.67\pm0.79$ & $69.33\pm0.34$ & $86.42\pm1.45$ & $56.03\pm0.27$ & $70.83\pm0.25$ & 8\\
    AD-GCL & $73.91\pm0.77$ & $73.28\pm0.46$ & $75.79\pm0.87$ & $88.74\pm1.85$ & $72.02\pm0.56$ & $90.07\pm0.85$ & $54.33\pm0.32$ & $70.21\pm0.68$ & 8.5\\
    SimGRACE & $79.12\pm0.44$ & $75.35\pm0.09$ & $77.44\pm1.11$ & $89.01\pm1.31$ & $71.72\pm0.82$ & $89.51\pm0.89$ & $55.91\pm0.34$ & $71.30\pm0.77$ & 5.6\\
    RGCL & $78.14\pm1.08$ & $75.03\pm0.43$ & $\underline{\textbf{78.86}\pm\textbf{0.48}}$ & $87.66\pm1.01$ & $70.92\pm0.65$ & $90.34\pm0.58$ & $56.38\pm0.40$ & $71.85\pm0.84$ & 5.1\\   
    LaGraph & $79.9\pm0.5$ & $75.2\pm0.4$ & $78.1\pm0.4$ & $\underline{\textbf{90.2}\pm\textbf{1.1}}$ & $77.6\pm0.2$ & $90.4\pm0.8$ & $56.4\pm0.4$ & $73.7\pm0.9$ & 3\\
    \midrule
    \name-MSE & $80.39\pm0.62$ & $75.58\pm0.23$ & $78.32\pm0.77$ & $88.04\pm1.56$ & $77.64\pm0.26$ & $90.26\pm0.80$ & $56.38\pm0.60$ & $\underline{\textbf{74.34}\pm\textbf{0.72}}$ & 2.8\\
    \name & $\underline{\textbf{81.12}\pm\textbf{0.40}}$ & $\underline{\textbf{75.92}\pm\textbf{0.15}}$ & $78.32\pm1.07$ & $89.19\pm1.83$ & $\underline{\textbf{78.90}\pm\textbf{0.32}}$ & $\underline{\textbf{91.08}\pm\textbf{0.56}}$ & $\underline{\textbf{56.71}\pm\textbf{0.30}}$ & $74.28\pm0.76$ & \underline{\textbf{1.5}}\\

    \bottomrule
  \end{tabular}
\end{table*}

\begin{table*}[t]
  \caption{Transfer learning on downstream graph classification tasks. We compared all the methods w.r.t. the AUROC score(\%). \underline{Bold} indicates the best performance on each dataset. - indicates that results are not available in published papers. Results for SOTA methods are as published.}
  \setlength{\tabcolsep}{4pt}
  \begin{tabular}{ccccccccc|c}
    
    \toprule
    Dataset & BBBP & Tox21 & ToxCast & SIDER & ClinTox & MUV & HIV & BACE & AVG.\\
    \midrule
    No Pre-Train & $65.8\pm4.5$ & $74.0\pm0.8$ & $63.4\pm0.6$ & $57.3\pm1.6$ & $58.0\pm4.4$ & $71.8\pm2.5$ & $75.3\pm1.9$ & $70.1\pm5.4$ & 66.96\\
    DGI & $68.8\pm0.8$ & $75.3\pm0.5$ & $62.7\pm0.4$ & $58.4\pm0.8$ & $69.9\pm3.0$ & $75.3\pm2.5$ & $76.0\pm0.7$ & $75.9\pm1.6$ & 70.29\\
    EdgePred & $67.3\pm0.24$ & $76.0\pm0.6$ & $64.1\pm0.6$ & $60.4\pm0.7$ & $64.1\pm3.7$ & $74.1\pm2.1$ & $76.3\pm1.0$ & $79.9\pm0.9$ & 70.28\\
    AttrMasking & $64.3\pm2.8$ & $\underline{\textbf{76.7}\pm\textbf{0.4}}$ & $64.2\pm0.5$ & $61.0\pm0.7$ & $71.8\pm4.1$ & $74.7\pm1.4$ & $77.2\pm1.1$ & $79.3\pm1.6$ & 71.15\\
    ContextPred & $68.0\pm2.0$ & $75.7\pm0.7$ & $63.9\pm0.6$ & $60.9\pm0.6$ & $65.9\pm3.8$ & $75.8\pm1.7$ & $77.3\pm1.0$ & $79.6\pm1.2$ & 70.89\\
    GraphCL & $69.68\pm0.67$ & $73.87\pm0.66$ & $62.40\pm0.57$ & $60.53\pm0.88$ & $75.99\pm2.65$ & $69.80\pm2.66$ & $\underline{\textbf{78.47}\pm\textbf{1.22}}$ & $75.38\pm1.44$ & 70.77\\
    JOAO & $70.22\pm0.98$ & $74.98\pm0.29$ & $62.94\pm0.48$ & $59.97\pm0.79$ & $81.32\pm2.49$ & $71.66\pm1.43$ & $76.73\pm1.23$ & $77.34\pm0.48$ & 71.9\\
    JOAOv2 & $71.39\pm0.92$ & $74.27\pm0.62$ & $63.16\pm0.45$ & $60.49\pm0.74$ & $80.97\pm1.64$ & $73.67\pm1.00$ & $77.51\pm1.17$ & $75.49\pm1.27$ & 72.12\\
    AD-GCL & $68.26\pm1.03$ & $73.56\pm0.72$ & $63.10\pm0.66$ & $59.24\pm0.86$ & $77.63\pm4.21$ & $74.94\pm2.54$ & $75.45\pm1.28$ & $75.02\pm1.88$ & 70.90\\
    SimGRACE & $71.25\pm0.86$ & - & $63.36\pm0.52$ & $60.59\pm0.96$ & -  & - & - & - & -\\
    GraphLoG & $71.04\pm1.86$ & $74.65\pm0.60$ & $62.32\pm0.51$ & $57.86\pm1.44$ & $78.72\pm2.58$ & $74.95\pm1.96$ & $75.12\pm1.98$ & $\underline{\textbf{82.6}\pm\textbf{1.25}}$ & 72.16\\
    RGCL & $71.42\pm0.66$ & $75.20\pm0.34$ & $63.33\pm0.17$ & $\underline{\textbf{61.38}\pm\textbf{0.61}}$ & $\underline{\textbf{83.38}\pm\textbf{0.91}}$ & $\underline{\textbf{76.66}\pm\textbf{0.99}}$ & $77.90\pm0.80$ & $76.03\pm0.77$ & 73.16\\
    \midrule
    \name & $\underline{\textbf{72.16}\pm\textbf{1.42}}$ & $75.58\pm0.56$ & $\underline{\textbf{64.27}\pm\textbf{0.23}}$ & $60.49\pm0.41$ & $82.29\pm1.21$ & $76.31\pm0.71$ & $77.08\pm0.83$ & $78.82\pm1.38$ & \underline{\textbf{73.37}}\\

    \bottomrule
  \end{tabular}
  \label{Table 4}
\end{table*}

\begin{table*}[t]
  \caption{Ablation study for \name~ on unsupervised learning datasets.}
  \setlength{\tabcolsep}{4pt}
  \label{Table 5}
  \begin{tabular}{ccccccccc|c}
    \toprule
    Dataset & NCI1 & PROTEINS & DD & MUTAG & COLLAB & RDT-B & RDT-M5K & IMDB-B & AVG.\\
    \midrule
    \name\_rd & $80.44\pm0.43$ & $74.90\pm0.70$ & $77.60\pm0.56$ & $86.36\pm1.93$ & $77.60\pm0.32$ & $90.02\pm0.96$ & $56.18\pm0.66$ & $73.22\pm0.77$ & 77.04\\
    
    \name\_nm & $80.79\pm0.31$ & $74.98\pm0.73$ & $77.39\pm0.33$ & $87.66\pm0.95$ & $77.46\pm0.37$ & $89.89\pm0.65$ & $56.17\pm0.56$ & $73.66\pm0.54$ & 77.25\\ 
    
    \name\_nB & $79.61\pm0.36$ & $75.19\pm0.36$ & $77.27\pm0.84$ & $86.68\pm1.94$ & $77.24\pm0.30$ & $89.80\pm0.48$ & $56.12\pm0.34$ & $73.66\pm0.84$ & 76.95\\ 
    
    
    \name & $\underline{\textbf{81.12}\pm\textbf{0.40}}$ & $\underline{\textbf{75.92}\pm\textbf{0.15}}$ & $\underline{\textbf{78.32}\pm\textbf{1.07}}$ & $\underline{\textbf{89.19}\pm\textbf{1.83}}$ & $\underline{\textbf{78.90}\pm\textbf{0.32}}$ & $\underline{\textbf{91.08}\pm\textbf{0.56}}$ & $\underline{\textbf{56.71}\pm\textbf{0.30}}$ & $\underline{\textbf{74.28}\pm\textbf{0.76}}$ & $\underline{\textbf{78.19}}$\\

    \bottomrule
  \end{tabular}
\end{table*}

\subsection{Transfer learning (RQ2)}

In transfer learning, we first pre-train a backbone model on Zinc-2M, and fine-tune the model on 8 multi-task binary classification datasets. All the results w.r.t. the area under receiver operating characteristic (AUROC) on downstream tasks are presented in Table \ref{Table 4},
as well as the average scores. 
From the table,
we see that
\name\ achieves the highest average score compared with other baselines, 
and also the best performances on the BBBP and Toxcast datasets. 
For other cases where \name\ is not the winner,
the gap between \name's score
and the winner's is small.
For example, the gaps between \name~ and the winner on the SIDER and MUV datasets are only 0.89\% and 0.35\%, 
respectively.
Further, let us take a closer look at GraphCL, which uses the same augmentation functions as \name. Specifically, \name~ leads GraphCL by $>6\%$ on both ClinTox and MUV datasets, 
and has an average score on all datasets that is 2.6\% higher than GraphCL.


\begin{figure*}[t]
  \centering
  \includegraphics[width=0.85\linewidth]{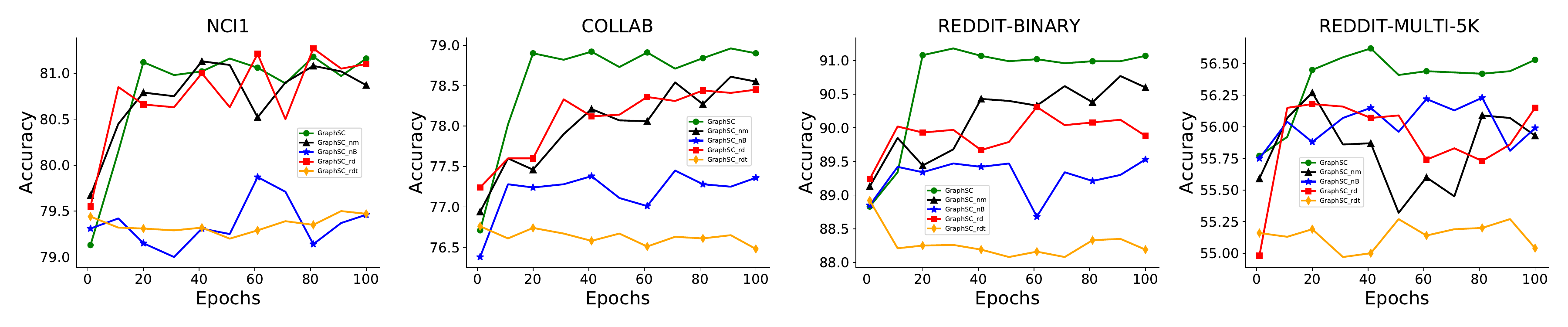}
  \caption{Convergence curves for \name~ and its variants on NCI1 , COLLAB, REDDIT-BINARY and REDDIT-MULTI-5K datasets}
  \label{Figure 3}
\end{figure*}

\begin{figure*}[t]
  \centering
  \includegraphics[width=\linewidth]{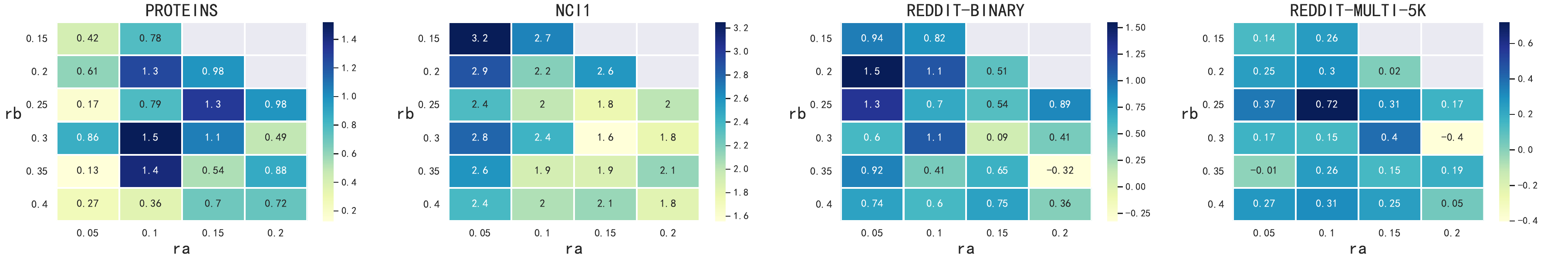}
  \caption{Accuracy difference between \name~ and GraphCL  versus various perturbation intensity pairs in unsupervised learning settings. 
  Since $r_{a}<r_{b}$, we fill in the parts of the diagram that violate this condition with blanks.}
  \label{Figure 4}
\end{figure*}

\begin{figure*}[!t]
  \centering
  \includegraphics[width=0.8\linewidth]{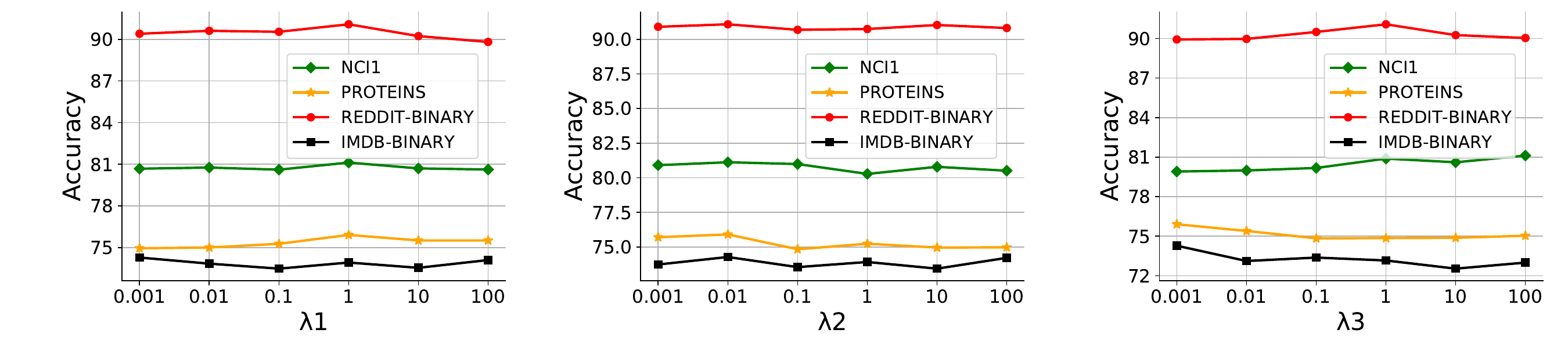}
  \caption{\name~’s sensitivity analysis w.r.t. the hyperparameters $\lambda_{1}$, $\lambda_{2}$ and $\lambda_{3}$ in unsupervised learning settings.
}
  \label{Figure 5}
\end{figure*}

\subsection{Ablation Study (RQ3)}
\label{AS}
We conduct an ablation study on \name~ to understand the characteristics of its main components. One variant randomly selects an augmented view of another sample as a negative sample. 
This is different from \name, which uses a negative sample that is directly constructed from the graph sample itself (see Section \ref{DA}). 
We call this variant \textbf{\name\_rd} (\textbf{r}an\textbf{d}om), which helps us evaluate the validity of our negative generation strategy.
Another variant trains model without considering masked self-contrast. 
This helps us understand the importance of masked self-contrast. 
We call this variant \textbf{\name\_nm} (\textbf{n}o \textbf{m}asked self-contrast).
Finally, to show the importance of the Barlow Twins loss regularization term, We remove $\mathcal{L}_{ab}$ from the objective function and call this variant \textbf{\name\_nB} (\textbf{n}o \textbf{B}arlow Twins). 
The results are given in Table \ref{Table 5} and we observe:

(1) \name~ achieves better performance than \name\_rd. Since \name\_rd randomly selects an augmented view of other sample as a negative, the performance gaps between \name~ and \name\_rd show that \name’s negative generation strategy is very effective in {selecting} a valid negative sample to improve classification accuracy.

(2) \name~ clearly outperforms \name\_nm on all datasets.
\name\_nm, which removes $\mathcal{L}_{ma}$ from the objective function, does not perform masked self-contrast. 
This leads to significant performance degradation due to the ignorance of the fact that graph-structured data generally contains multiple aspects of information.

(3) \name~ beats \name\_nB across all datasets. In particular,
\name~significantly outperforms \name\_nB on MUTAG and
COLLAB. 
This shows that the Barlow Twins loss regularization is particularly important for model training. When using Barlow Twins loss regularization, the model can explicitly shorten the absolute distance between the anchor and the positive sample. 
This effectively compensates for the inherent weakness of triplet loss.


\subsection{Convergence analysis (RQ4)}
We next show how different components of \name~ 
can address the hard-to-train due to the usage of triplet loss objective.
To show the difficulty in training triplet loss based objective,
we further consider a variant of \name~that randomly selects an augmented view of another sample as a negative sample and uses the triplet loss to train GNN encoder. 
We call it \textbf{\name\_rdt} (\textbf{r}an\textbf{d}om selection and \textbf{t}riplet loss).
We take it as a reference and 
show the convergence results of \name\ variants 
on four datasets 
in Fig. \ref{Figure 3}. 
From the figure, 
we observe that:

(1) The accuracy of \name\_rdt does not increase with more training epochs on all four datasets.
This shows that \name\_rdt 
is hard-to-train due to the triplet loss objective.

(2) \name~ converges faster than \name\_nm. 
Specifically, on the COLLAB and REDDIT-BINARY datasets, \name\_nm converges with more than 60 epochs, 
while \name~ uses only 20 epochs. In addition, \name\_nm trains very unsteadily on REDDIT-MULTI-5K.
The convergence speed gaps between \name~ and \name\_nm show that \name's masked self-contrast is very effective in accelerating model convergence.

(3) Compared with \name,
\name\_nB converges slower and performs worse. 
\name\_nB, 
which only consider relative distances between anchor and positive sample, might not pull close positive sample pairs well. 
By explicitly shortening the absolute distance for positive sample pairs,
\name\ converges faster and performs better.

(4) \name~ also achieves faster convergence 
than \name\_rd. 
This shows that the negative generation strategy is particularly important for model training. 
When using our proposed negative generation strategy, 
the model can obtain informative negative samples, which accelerates convergence.

\subsection{Hyperparameter sensitivity analysis (RQ5)}
We end this section with a sensitivity analysis on the hyperparameters of \name. In particular, we study four key hyperparameters:
the perturbation intensities \begin{math} r_{a},r_{b} \end{math}, and the term weights $\lambda_{1}$, $\lambda_{2}$, $\lambda_{3}$.
We evaluate the performance of \name\ on unsupervised settings.
We vary one hyperparameter each time with others fixed.
Experimental results are given in Fig.~\ref{Figure 4} and Fig.~\ref{Figure 5}.

\noindent{\small$\bullet$}
{\textbf{Perturbation intensity}}: 
As shown in Fig. \ref{Figure 4}, we calculate the accuracy difference between \name~and GraphCL for different combinations of perturbation strengths. 
Although the best settings of perturbations are different for each dataset, 
\name\ performs very well over various perturbation strength combinations.
This demonstrates the stability of the model.
 
\noindent{\small$\bullet$}
{\textbf{ Term weight $\lambda_{1}$, $\lambda_{2}$ and $\lambda_{3}$}}:
As can be observed in Fig. \ref{Figure 5}, for the term weight $\lambda_{1}$, $\lambda_{2}$ and $\lambda_{3}$, \name~ gives very stable performances over a wide range of parameter values. 
This also shows the robust performance of \name.


 

\section{CONCLUSIONS}
In this paper, we studied graph-level representation learning and proposed \name, a graph contrastive learning framework, which uses triplet loss as objective. 
\name~ first uses graph augmentation functions of different intensities to obtain a positive and negative view of a graph sample from the graph itself. 
Further, 
it factorizes the graph representations into multiple factors
and then presented a self-contrast mechanism to separate positive and negative samples.
Moreover, \name\ tries to shorten the absolute distance between an anchor and its positive sample, which addresses the problem of triplet loss in optimizing only the relative distance between the anchor point and its positive/negative samples.
We conducted extensive experiments in both unsupervised and transfer learning settings, and experimental results demonstrate the effectiveness and transferability of the proposed framework.

\bibliographystyle{IEEEtran}
\bibliography{IEEEexample}

\end{document}